# An Optimal Itinerary Generation in a Configuration Space of Large Intellectual Agent Groups with Linear Logic

Dmitry Maximov
Institute of Control Science
Russian Academy of Science
Moscow, Russia
dmmax@inbox.ru

*Abstract* — a group of intelligent agents which fulfill a set of tasks in parallel is represented first by the tensor multiplication of corresponding processes in a linear logic game category. An optimal itinerary in the configuration space of the group states is defined as a play with maximal total reward in the category. New moments also are: the reward is represented as a degree of certainty (visibility) of an agent goal, and the system goals are chosen by the greatest value corresponding to these processes in the system goal lattice.

## I. INTRODUCTION

The artificial intelligence is represented in the Artificial General Intelligence (AGI) approach as an information processor which consumes and gives out information. Investigations in the field are focused on systems which *act* rationally. A formal description of the most intelligent agent (AIXI) behavior, in the sense of some intelligence measure, is suggested in AGI framework [1]. The model is based on probabilistic modeling of the environment, and on the next system move determination based on previous experience, and on a numerical estimation of the system position reward and on the maximization of the expected reward along the trajectory. However, the method to obtain this numerical estimation is absent. Also, there are no models to describe such *agent groups'* behavior.

It has been demonstrated in [2-5] that the structure existence (a lattice structure or else a monoid structure, i.e. the linear logic structure) in the system task [3, 5] or goal [2] set is sufficient for the system to behave quite reasonable. The behavior looks even like ants' behavior in something [5]. But it is not supposed here the environment modeling unlike [1]. In this paper, the topic develops based on the idea that it is possible to represent different aim parallel achievement processes fulfilling by various intelligent agents in some environment as a tensor multiplication in linear logic. The logic is modeled in some game category [6]. Thus, it is possible to describe an itinerary choice of the goal achievement process by the intelligent agent system in some environment as a game. Position rewards in the game are represented by sets which describe the information about goals or their distinctness degree. Thus, the rewards are provided by the environment, and they are lattice elements but not numbers as in [1]. These lattice and linear logic structures are provided by the environment. But similar structures are also given by our ideas about the system and agent purposes. These are used to determine the priority of different parallel processes in such game category.

## II. MATHEMATICAL BACKGROUNDS

### A. Lattices[4,7]

*Definition 1*: A *partially-ordered* set P is the set with such a binary relation $x \leq y$ for elements in it, that for all $x, y, z \in P$ the following relationships are performed:
- $x \leq x$ (reflexivity);
- if $x \leq y$ and $y \leq x$, then $x = y$ (anti-symmetry);
- if $x \leq y$ and $y \leq z$, then $x \leq z$ (transitivity).

The definition means that in the partially-ordered set not all elements are compared with each other. This property distinguishes these sets from linear-ordered ones, i.e., from numeric sets which are ordered by a norm. Thus, the elements of the partially- ordered set are the objects of more general nature than numbers. In the partially-ordered set diagram, the greater the element (i.e., vertex, node) is the higher it lies, and the elements are compared with each other lie in the same path from the minor element to the greater one. Two examples of a partially-ordered set diagram are represented in Fig. 1 which are also lattice diagrams.

*Definition 2*: The *upper bound* of a subset X in a partially-ordered set P is the element $a \in P$, containing all $x \in X$. The supremum or *join* is the smallest subset X upper bound. The infimum or *meet* defines dually as the greatest element $a \in P$ containing in all $x \in X$.

*Definition 3*: A *lattice* is a partially-ordered set, in which every two elements have their meet, denoting as $x \wedge y$, and join, denoting as $x \vee y$.

In the lattice diagram the elements join is the nearest upper element to both of them, and the meet is the nearest lower one

The project was partly supported by RFBR grant 16-08-00832a to both. The elements generating by joins and meets all other elements are called *generators*.

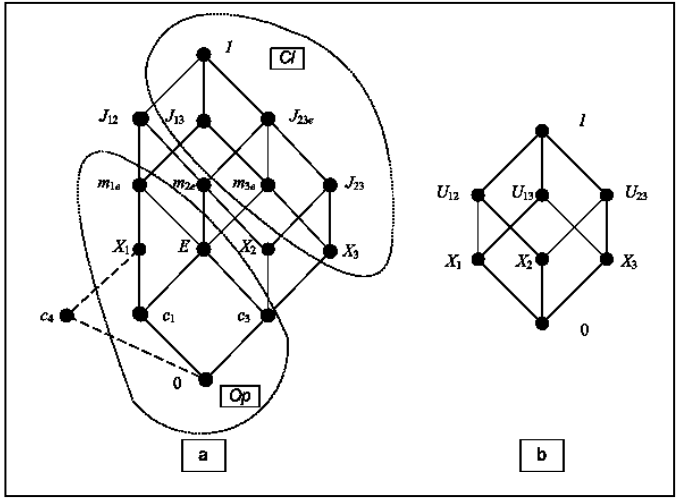

Fig. 1. Examples of a goal lattice (a) and of an agent desire lattice (b)

They refer to the lattice as *complete lattice* if its arbitrary subset has the join and the meet. Thus, any complete lattice has the greatest element "1", and the smallest one "0" and every finite lattice is complete.

### B. Linear Logic[4,8]

If a multiplication operation is additionally defined at the lattice elements, then the operations of linear logic also exist at the lattice. We use the phase semantic of linear logic from [8].

**Definition 4**: A *phase space* is a pare $(M, \perp)$, where $M$ is a multiplicative monoid (i.e., a triple $(M_0, \cdot, e)$ with $M_0$ is a set and $\cdot$ is a multiplication with the unit $e$), which is also a lattice and the element *false* of the lattice $\perp \subset M$ is an arbitrary subset of the monoid.

In linear logic, the element false differs from 0 (the minimal lattice element) in general in contrast to classical logic or intuitionistic one. The multiplication $X \cdot Y = \{x \cdot y | x \in X; y \in Y\}$ is defined for arbitrary monoid subsets (i.e., the lattice elements) $X, Y \subset M$. The linear implication $X \Rightarrow Y = \{z | x \cdot z \in Y, \forall x \in X\}$ is also defined. For $X \subset M$ its dual is defined as $X^\perp \Rightarrow \perp$. The dual element is a generalization of the negation in the case of linear logic.

**Definition 5**: *Facts* are such subsets $X \subset M$ that $X^{\perp\perp} = X$ or equivalently $X = Y^\perp$ for some $Y \subset M$.

Thus, facts are lattice elements coinciding with their double negations. E.g. $\perp^\perp = I = \{e\}^{\perp\perp}$; $1 = M = \emptyset^\perp$; $0 = 1^\perp = M^\perp = \emptyset^{\perp\perp}$. Here 1 is the maximal element of the lattice $M$, 0 is its minimal element, $e$ is the monoidal unit, and $I$ is the neutral element of the multiplicative conjunction (see after this).

It is easy to get the next properties: $X^\perp X \subset \perp$; $X \subset X^{\perp\perp}$; $X^{\perp\perp\perp} = X$; $X \Rightarrow Y^\perp = (X \cdot Y)^\perp$; $(X \vee Y)^\perp = X^\perp \wedge Y^\perp$. From here we get only facts may be the values and the consequents of the implication.

At facts the lattice operations of the additive conjunction & and the additive disjunction + are defined in the following way:
$X \& Y = X \wedge Y = (X^\perp \vee Y^\perp)^\perp$; $X + Y = (X^\perp \& Y^\perp)^\perp = (X^\perp \wedge Y^\perp)^\perp = (X \vee Y)^{\perp\perp}$. The duality of the operations understands here as in the set theory: $\vee^\perp = \wedge$ $\wedge^\perp = \vee$ in which the duality means the negation.

At facts, multiplicative operations are also defined. These are the multiplicative conjunction $\otimes$ and the multiplicative disjunction ð: $X \otimes Y = (X \cdot Y)^{\perp\perp} = (X \Rightarrow Y^\perp)^\perp = (X^\perp ð Y^\perp)^\perp$; $X ð Y = (X^\perp \cdot Y^\perp)^\perp = X^\perp \Rightarrow Y$. The neutral element of the operation & is 1, the dual to it (neutral element of the operation +) is 0. The neutral element of the operation ð is $\perp$, the dual to it, the neutral element of the operation $\otimes$ is $I$.

The set of facts is divided into two classes dual to each other: the class of *open* facts **Op** and the class of *closed* facts **Cl**. The set **Op** is closed by operations + and $\otimes$. Its maximal element by inclusion is $I$, and the minimal one is 0. The set **Cl** is closed correspondingly by operations & and ð, and its maximal element is 1, and the minimal one is $\perp$. In Fig. 1 a, the class of *open* facts is encircled as **Op** with $I = m_{1e}$, and the class of *closed* facts is encircled as **Cl** with $\perp = X_3$

### C. Game Semantics in a Linear Logic Category[6]

**Definition 6**: A *Conway game* is defined as a rooted graph with vertices $V$ as the game positions and edges $E \subset V \times V$ as the game moves. Each edge has a polarity $\pm 1$ which depends on whether it is the Proponent or the Opponent move.

**Definition 7**: A *trajectory or a play* is some path from the graph root $*$. The path is *alternated* if the adjacent edges are of different polarities.

**Definition 8**: A *strategy* is defined as a non-empty set of alternated paths of even length, which are started from the Opponent move, closed up to the prefix of even length and determined. Determinism means that two paths with the common prefix, which differ in two moves, should coincide.

**Definition 9**: A *dual play* $X^\perp$ is obtained from the play $X$ by reversing the polarity of moves.

**Definition 10**: The tensor product $X \otimes Y$ of two Conway games $X$ and $Y$ is defined in such a way:

positions $x \otimes y$ are $V_{X \otimes Y} = V_X \times V_Y$ with the root

$*_{X \otimes Y} = *_X \times *_Y$, moves are $x \otimes y \mapsto \begin{cases} z \otimes y, & x \mapsto z \text{ in } X \\ x \otimes z, & y \mapsto z \text{ in } Y, \end{cases}$

and the polarity of a move in $X \otimes Y$ is inherited from the polarity of the underlying move in $X$ or $Y$.

A generalized linear logic is modeled in a category of such games. The category objects are Conway games and morphisms $X \to Y$ are strategies in $X^\perp ð Y$. These multiplications are linear implications $X^\perp ð Y = X \Rightarrow Y$. It should be mentioned that on game graphs the operations ð and $\otimes$ are the same so in [6] they are not even distinguished. A Conway game with a payoff is the play with an additional weight 1, 1/2 or 0 in each vertex. The weight depends on whether the position is winning (i.e., of the weight 1 or 1/2) or not. In the tensor product, these weights obey rules of Boolean conjunction and implication. A strategy is winning if it terminates in the winning position. In the category of Conway games with a payoff, morphisms are *winning* strategies now. It is possible to prove that the categorical construction is conserved if the weights' numbers are replaced with some sets form a lattice, and the Boolean operations are replaced with the lattice operations (different variants of the replacement in [9, 10]). The greater set is connected with a position, the more advantageous it is, and it is winning if its weight is not 0. We suppose the existence of a universal set containing all the others. Thus, all such estimation sets form a complete Brouwer lattice.

III. Behavior Determination of an Intellectual Agent System

An example of the goal lattice $M_s$ matches a system of $l$ agents (Fig. 1, a). In the lattice vertices, $X_i$ and $E$ are the generators and denote the system goals. Vertices $J_i$ are the generators joins and denote combined goals achieving. These combined goals, as well as generators, may be associated with the correspondent task fulfilling thus the vertices may have meets, i.e., some subtask included in different tasks (or correspondent goals). The higher goal lies (hence the more tasks it contains), the more important this behavior variant is. It supposed that the preferable behavior of the system is to achieve all its goals. This variant corresponds to the top lattice element 1. And the bottom element 0 corresponds to complete inactivity and to the least essential behavior variant. All the estimations may be considered as partially true truth values. Thus we can say that the more important the behavior is, the truer it is.

The agent lattice $M_i$ (Fig. 1, b) has another meaning. The same generators and their joins mean the agents' desires. One of the vertices is marked as an intention for every agent. In the same manner, the more desires are included in the intention, the more essential the intention is.

A variant of the open (*Op*) and the closed (*Cl*) classes definition is indicated in the system lattice (Fig. 1, a). This definition (as well as the multiplication definition) defines the structure of linear logic in the lattice. The element multiplication is obtained (usually ambiguously, [5]) from the demand to implement the linear logic operations properties. In this case, it is possible to consider parallel processes of combined goals achieving as the tensor product of corresponding lattice elements in the logic. And the priority of different processes is obtained from the demand of the greatest correspondent tensor product estimation in the lattice (the product is an element of the lattice; hence the higher it lies, the more important it is, i.e., the greater its truth-value is).

It is supposed some initial agent goal distribution. For such a system, we consider the process of the interaction of the system with the environment as a Conway game. In the game, the environment is the Proponent which provides the system (the Opponent) with the information about the environment objects. The Opponent moves from one position in the environment to the other by the use of the information to achieve his goals. For the system of $l$ agents we, in fact, have $l$ parallel processes and, therefore, the resulting game is their tensor product. Thus, agents are placed initially in the configuration space (environment) in the root $*=*_1 \otimes ... \otimes *_l$ of the system game $A = A_1 \otimes ... \otimes A_l$ with the system goal lattice $M_s$ and agent intention lattices $M_i$, $i = 1..l$.

The game $A_j$ represents *possible* agent moves in the environment. But the *real* trajectory or the play is chosen from the demand of the maximal total position reward along the projected path. The agent move in the environment is estimated corresponding to an optimality criterion with the reward $k(p_i, b_j)$ in the position $p_i$ of the goal $b_j$ achieving process.

It may be that the system does not see any goal initially and moves according to a criterion of an optimal move in goals absence. Thus, the system has additionally $l$ goals (tasks) $a_j$ of movement in the environment, which are included as in the generator set of the lattice $M_s$, as in the generator sets of lattices $M_i$.

The optimality criterion may represent the highest degree of the correspondence to the demanding system configuration, or the highest freedom in future moves or the most excellent visibility from a position or so on in the case of the free movement in the goals absence. And in the case of the system goals achieving we suppose the better a goal object is visible, the higher the reward is.

It is supposed that agents see not the whole environment, but up to some horizon which may change in each direction Fig. 2. Moreover, the agent sees the environment as in the fog – the nearer the object, the better it is perceived[1]. Therefore, we can predict a play only up to some finite step number with the increasing uncertainty along the path.

Let us $n$ goals $b_1...b_n$ are discovered in the environment by the system with information about them (or some other reward) $k(p_i, b_j)$ in positions $p_i$ of the play $A = A_1 \otimes ... \otimes A_l$ of $l$ agents. The game $A$ corresponds to $l$ parallel processes of achieving $l$ movement goals $a_j$. Then, a winning strategy of the game $A' = A_1 \otimes ... \otimes A_l \Rightarrow B_1 \otimes ... \otimes B_k$ defines a transition (morphism) to this new game $A'$. This game corresponds to $l$ parallel processes of moving and achieving those $k$ goals from discovered $n$ ones, which may be better achieved in the next sense (that means that these $k$ goals are the system intentions).

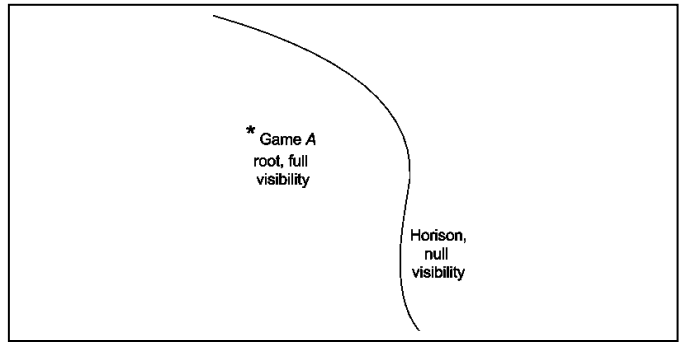

Fig. 2. The visibility horizon

---

[1] It is possible to consider different horizons and different uncertainty degree for different goals. We do not do it here.

The reward $k(p_i, b_j)$ is some set. In the case of, e.g., unmanned vehicles, it may be an image of the object. The better the object is visible, the better the coincidence of the image with the original and the greater the reward.

It is reasonable to choose the trajectory (play) from the demand to maximize the reward along the path within the visibility horizon:

$$k_{play}^{(A_1 \otimes ... \otimes A_l)^\perp ð B_1 \otimes ... \otimes B_k} = \quad (1)$$

$$= max_{plays} \left[ \bigcup_{play} k^{(A_1 \otimes ... \otimes A_l)^\perp} \bigcup_{play} (k^{B_1} \& ... \& k^{B_k}) \right]$$

Here the reward $k_{play}^{(A_1 \otimes ... \otimes A_l)^\perp ð B_1 \otimes ... \otimes B_k}$ is maximized in the game $A'$ and corresponds to that process of $k$ goals achieving that has the highest priority $(a_1 \otimes ... \otimes a_l)^\perp ð b_1 \otimes ... \otimes b_k$ in the system goal lattice. The priority is maximal among all possible parallel processes of achieving $n$ discovered goals. The maximum is taken among all possible plays and it joins the rewards along these plays (i.e., trajectories) in $(A_1 \otimes ... \otimes A_l)^\perp$, and in $B_1$, and so on and in $B_k$. Thus, the sign & means conjunction.

If there are parallel processes which are not compared by priority, i.e., if there are several incompatible priorities $(a_1 \otimes ... \otimes a_l)^\perp ð b_1 \otimes ... \otimes b_k$, it is possible to reorder the goal lattice in the manner that some lattice vertex has used as an additional priority [3]. In the reordered lattice, initially incompatible elements may become compatible, so we can choose the processes' priority. Such a lexicographical rule follows from the teleological system assignment, i.e., our vision of the system purpose. It may be an occasion that the lattice is so symmetric that does not allow make such a choice. In this case, the ambiguity of the linear logic structure allows the necessary variant choosing [9].

Also, if we are not able to determine the trajectory in (1) uniquely, it is possible to use the goal lattices of different agents to determine the agent (and the correspondent trajectory) priority as in [2]. In this case, we assign a definite agent to achieve a particular goal from agents' intention estimations.

Specifically, let us there are three agents' plays $A_1, A_2$ and $A_3$ in which the agents in positions $p_1$ have discovered two goals $b_1$ and $b_2$ from possible three ones (Fig. 3).

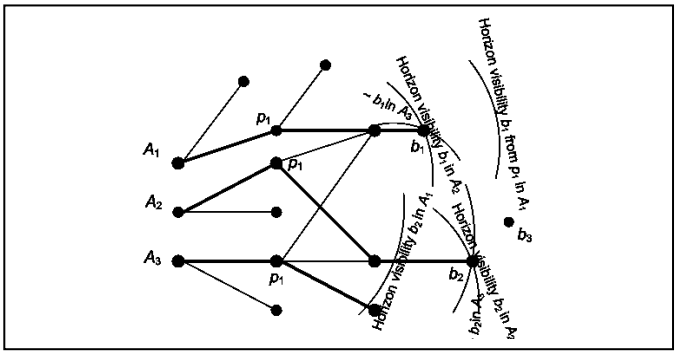

Fig. 3. An example of a game (depicted only the Opponent moves)

Let us also the rewards in the positions $p_1$ satisfy the next relations:

$$k_{A_1}(p_1, b_1) > k_{A_2}(p_1, b_1) \gtrless k_{A_3}(p_1, b_1) \quad (2)$$

$$k_{A_3}(p_1, b_2) > k_{A_2}(p_1, b_2) \gtrless k_{A_1}(p_1, b_2) \quad (3)$$

where the sign $\gtrless$ means the rewards' incompatibility. This means that though the visibility degrees in positions $p_1$ of $A_2$ and $A_3$ does not differ, the foreshortenings are different thus the images are different and, therefore, incompatible. Therefore, we cannot define the system trajectory uniquely.

It is clear that agent-1 should achieve the goal $b_1$ because it sees it better than the others. For the others let us, in this case, the desires' lattices for these two agents have the form of Fig. 4. Thus, the agent-2 may reach goals $b_1$ and $b_2$, and the agent-3 has an additional desire $b_3$. Then, we may evaluate the lattice vertex weights by the formula [2]:

*Vertex weight =*

*= joined in the vertex desires /total desires' number* (4)

The formula evaluates vertices' weights by two parameters: the number of desires joined in the intention and the vicinity of the vertex to the top element, i.e., to the most desirable behavior variant. The play of the two parameters allows weights' comparing in different lattices. Thus, e.g. the $U_{12}$ weight is $V_{U_{12}} = 2/3$. Since $V_{b_2} = 1/2$ in the agent-2 lattice and $V_{b_2} = 1/3$ in the agent-3 lattice, we should choose the bigger value and correspondent agent-2 to achieve the goal $b_2$. This is reasonable because the agent-3 is left free and can look for the goal $b_3$. Finally, the obtained play is depicted in Fig. 4 by bold lines.

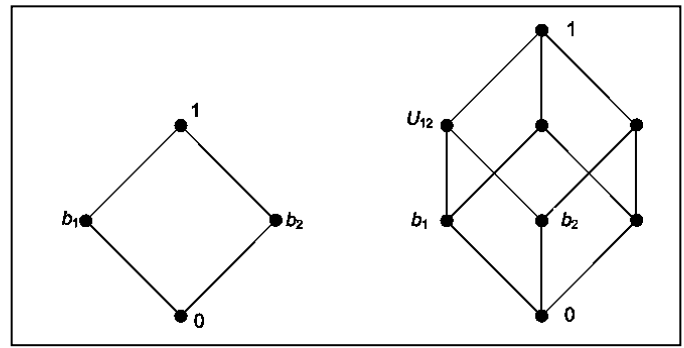

Fig. 4. The agent-2 lattice (a) and the agent-3 lattice (b)

The additional subtle point is that the graphs of the games $A$ and $B$ do not always coincide, i.e., a goal may be seen in $B$, but the way to the goal may not exist in $A$ (i.e., in the environment). Thus, we have a deal with the game $A$ of $l$ parallel processes of agent moves, on the positions of which the rewards of $k$ games $B$ are possible but are not obligatory.

It should be pointed out that the structures of tensorial multiplications $\otimes$ and $ð$ in $(a_1 \otimes ... \otimes a_l)^\perp ð b_1 \otimes ... \otimes b_k$ and in $(A_1 \otimes ... \otimes A_l)^\perp ð B_1 \otimes ... \otimes B_k$ are different. In

$(a_1 \otimes ... \otimes a_l)^\perp ð b_1 \otimes ... \otimes b_k$ the tensor $\otimes$ is a monoid in the goal lattice. These structures of ð and $\otimes$ are pre-existent and does not depend on the environment. It may be chosen from very general considerations [4] and determines the system purpose. But the tensor $\otimes$ and the co-tensor ð in $(A_1 \otimes ... \otimes A_l)^\perp ð B_1 \otimes ... \otimes B_k$ are determined by the environment, by the goals and obstacles distributions, by visibility and so on. Thus, there are two different linear logic structures in the approach.

## IV. Conclusion

In the paper, an itinerary choice for a movement of an agent group with its goal lattice was considered in some environment. The original method is valid for large intelligent agent groups and it is of great importance for optimal large agent groups moves since it uses two types of reward estimations. Each agent move was represented as a game in which the environment corresponds to the Proponent which provides the Opponent (the agent) with some information to achieve the agent goal. Different agent moves are considered as parallel processes which are represented as the tensor product of correspondent games and form a comprehensive game.

The game has rewards on its positions, which estimates the quality of the information provided by the environment. The reward may be very different: it may represent the highest degree of the correspondence to the demanding system configuration, or the highest freedom in future moves, or the most excellent visibility from a position, or the degree of the coincidence of the goal object image with the original or some else. We demand the greatest total reward along all agents' plays to choose the group trajectory. And we choose those goal achieving processes (i.e., agent plays) from all possible, which have the highest estimation in the agent group goal lattice. It is so because every goal and the corresponding process of its achieving has a definite correspondent truth value in the lattice. The higher the value lies in the lattice diagram, the higher priority of the process is.

Thus, we consider two types of estimations: the goal lattice value defines the choice of the goal achieving processes from all possible ones, and the position rewards of the game determine the optimal trajectory of these chosen processes in the environment. When it is not possible to determine the most significant estimations in both these cases, some additional methods may be used to select the optimal variant.